\documentclass[letterpaper, 10 pt, conference]{ieeeconf}  

\IEEEoverridecommandlockouts                              

\usepackage{graphics} 
\usepackage{amsmath} 
\usepackage{graphicx}
\usepackage{multirow}
\usepackage{makecell}
\usepackage{xcolor}

\usepackage{hyperref}
\hypersetup{
    colorlinks=true,
    linkcolor=black,
    filecolor=magenta,      
    urlcolor=black,
    pdftitle={Overleaf Example},
    pdfpagemode=FullScreen,
    }

\title{\LARGE \bf
OmniRace: 6D Hand Pose Estimation for Intuitive Guidance of  

Racing Drone
}

\author{Valerii Serpiva, Aleksey Fedoseev, Sausar Karaf, Ali Alridha Abdulkarim, Dzmitry Tsetserukou
\thanks{The authors are with the Intelligent Space Robotics Laboratory, Skolkovo Institute of Science and Technology Moscow, Bolshoy Boulevard 30, bld. 1, 121205, Moscow, Russia. 
\tt \{Valerii.Serpiva, Aleksey.Fedoseev, Sausar.Karaf, Ali.Abdulkarim, D.Tsetserukou\}@skoltech.ru}
}

\begin{document}

\maketitle
\thispagestyle{empty}
\pagestyle{empty}

\begin{abstract}

This paper presents the OmniRace approach to controlling a racing drone with 6-degree of freedom (DoF) hand pose estimation and gesture recognition. To our knowledge, this is the first technology enabling low-level control of high-speed drones through gestures. OmniRace employs a gesture interface based on computer vision and a deep neural network to estimate 6-DoF hand pose. The advanced machine learning algorithm robustly interprets human gestures, allowing users to control drone motion intuitively. Real-time control tests validate the system's effectiveness and its potential to revolutionize drone racing and other applications. Experimental results conducted in simulation environment revealed that OmniRace allows the users to complite the UAV race track significantly (by 25.1\%) faster and to decrease the length of the test drone path (from 102.9 to 83.7 m). Users preferred the gesture interface for attractiveness (1.57 UEQ score), hedonic quality (1.56 UEQ score), and lower perceived temporal demand (32.0 score in NASA-TLX), while noting the high efficiency (0.75 UEQ score) and low physical demand (19.0 score in NASA-TLX) of the baseline remote controller. The deep neural network attains an average accuracy of 99.75\% when applied to both normalized datasets and raw datasets. OmniRace can potentially change the way humans interact with and navigate racing drones in dynamic and complex environments. The source code is available at https://github.com/SerValera/OmniRace.git.

\end{abstract}

\section{Introduction}

\begin{figure}[t]
\centerline{\includegraphics[width=0.4875\textwidth]{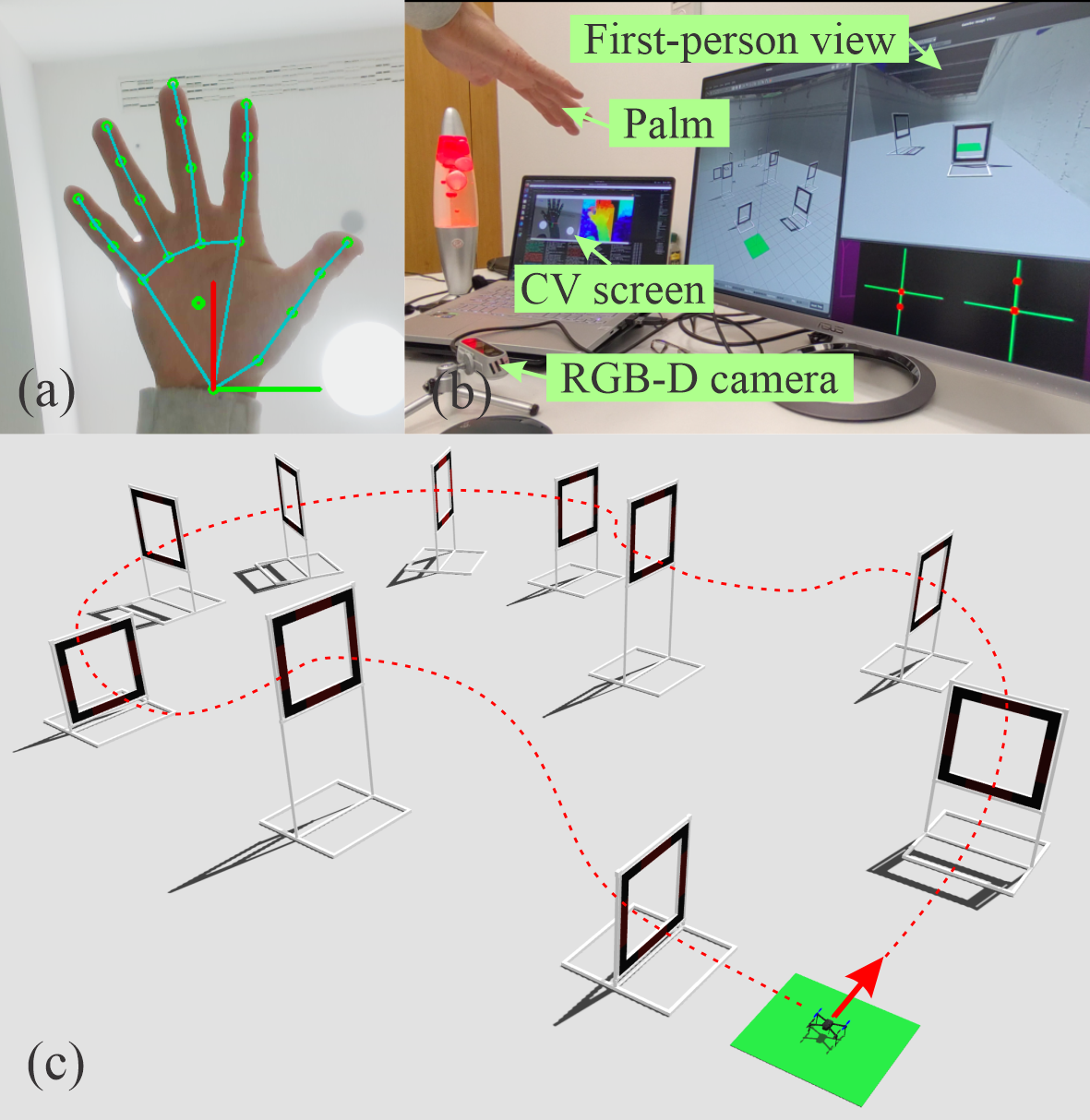}}
\caption{(a) Visual estimation of human hand pose in 6-DoF and gesture recognition. (b) Experimental setup for evaluating the gesture control interface. (c) Drone race track with 10 gates in a simulation environment used for evaluating the control interfaces.}
\label{fig:teaser}
\end{figure}

Drone racing, particularly from a first-person perspective, has been extensively explored due to the unique challenges it poses, such as adaptive control, navigation in cluttered environment, and human decision-making \cite{human_drone_race}. In this scenario, pilots navigate the race track via a video feed from an onboard camera. The drone's ability to move with six degrees of freedom within three-dimensional space requires pilots to have advanced visual-motor coordination skills to successfully complete the race within the shortest possible time. The difficulty in controlling drones is caused by a multitude of factors, including the high velocity of the drone's flight, the requirement for precise control to avoid obstacles, and the cognitive ability to process visual feedback from the drone’s perspective. Learning to skillfully control a racing drone is not a simple task. It demands extensive practice over a prolonged period. However, the duration varies significantly among individuals, with estimates ranging from several weeks to months, depending on the frequency of practice, as well as individual's inherent motor abilities \cite{dr_learn}.

Controlling drones through hand motion offers an immersive and intuitive approach within the field of robotics. This method emulates a natural human-machine interface, allowing individuals to navigate robots in a manner akin to the manipulation of physical objects. While this technique may have some limitations compared to the precise control of traditional systems, it opens up a fascinating area of research that explores the integration of human motor skills with robotic control. This form of interaction is reminiscent of the Jedi's use of “The Force” in the popular Star Wars franchise \cite{Rein2023}, a notion that has captivated audiences worldwide. The allure lies in the seemingly magical ability to influence one's surroundings through mere gestures, creating a sense of empowerment and control. This paper aims to explore this fascinating concept and investigate the potential of using this "force" to control racing drones through human hand movements. We explored how this innovative technology works, its potential applications, and the future it holds for drone racing.

In this work, we present a system for controlling a drone using just gestures, without any additional wearable devices or controllers. We build on our previous work, SwarmPaint \cite{swarmpaint}, where we introduced the concept of a swarm control interface that allows an operator to guide the swarm through path drawing and formation control. Our current research presents an upgraded version of the drone guiding interface. In addition to high-level commands, the OmniRace interface allows for low-level control of the drone's position by setting the orientation and throttle for the drone's autopilot. This enables the user to understand the current orientation of the drone and perform more complex maneuvers with just hand movements. Additionally, we tested our gesture interface and compared it against a baseline remote controller on a racing track in a simulation environment. 

\section{Related Work}

\subsection{Human-Robot Interaction}

The field of Human-Robot Interaction (HRI) has seen significant advancements in recent years \cite{State_of_the_art}, particularly in the area of gesture-based control \cite{gest_based}. Cutting-edge studies have been conducted, resulting in the development of cognitive human-machine interfaces and interactions that support adaptive automation in one-to-many applications, where a single human operator monitors and coordinates multiple Unmanned Aerial Vehicles (UAV) \cite{hri_adaptive}. User-case studies conducted in the field of HRI demonstrate that drones are being actively utilized in real world scenarios. For instance, HRI research is aiming at medicinal delivery \cite{hri_med} or forest firefighting missions \cite{hri_fire}. This necessitates a working environment where humans and robots coexist and have the capability to effectively control and interact with drones. The human-drone interaction aspect of this research is crucial to ensure these technologies are user-friendly, safe, and efficient in real-world scenarios. For example, HyperPalm \cite{nazarova2022hyperpalm} is a novel gesture interface designed for intuitive interaction with quadruped robots. DroneARchery \cite{dorzhieva2022dronearchery} explores augmented reality-driven human-drone interaction using RL-based swarm behavior. However, challenges remain, such as improving the robustness of gesture recognition in diverse environments \cite{Challenges_Industrial}, reducing the computational requirements of machine learning (ML) and artificial intelligence (AI) algorithms \cite{ai_challenges}, and enhancing the safety and reliability of gesture-based drone control \cite{safety}. 

\subsection{Sensors and Devices for Gesture Detection}

The technology relies on sensors and devices that can detect human hand movements and gestures. These devices include but are not limited to IMU, depth cameras, and infrared sensors. Accelerometers and gyroscopes \cite{imu_gesture} are often embedded in wearable devices to detect hand motion and to tackle muscle sensors \cite{sensor_muscle}, while depth cameras \cite{depth_gesture} and infrared sensors \cite{ir_gesture} are used to capture hand shape and gestures in 3D. Also the combination of the sensor can provide advance control of the swarm of drones \cite{swarm_glove}. These sensors provide raw data which is then processed and interpreted as control commands for the drone. For instance, a specific hand gesture could be interpreted as a command to move the drone forward, backward, left, or right \cite{high_level}. Wearable devices provide a large amount of data, however, they reduce mobility and ease of control. Visual-based sensors \cite{camera_gesture} are most actively used in human hand gesture detection for several reasons: they offer high accuracy, are non-intrusive, allow for real-time interaction, and can be integrated with Machine Learning and Artificial Intelligence. Most of the proposed control methods provide the opportunity to control a drone at a high level of commands, which does not allow for quick maneuvering in conditions of drone racing at high speeds.

\subsection{ML and AI Methods for Gesture Recognition}

ML and AI play crucial roles in the interpretation of sensor data into meaningful commands. ML algorithms are used to train models that can recognize specific hand gestures based on sensor data \cite{ml}. These algorithms can learn from a set of training data containing various hand gestures and their corresponding drone commands, allowing them to accurately identify these gestures in real-time. On the other hand, AI techniques such as neural networks \cite{nn} and deep learning \cite{dl} are used to improve the accuracy and efficiency of gesture recognition. These techniques allow the system to learn complex patterns and relationships between different hand gestures and drone commands, enabling it to handle a wider range of gestures and more complex control tasks. Despite these advancements, there are still challenges to be addressed. These include the variability in human gestures due to differences in individual style, speed, and amplitude of gestures, as well as environmental factors such as lighting conditions and background noise \cite{challenges}. Furthermore, the computational complexity \cite{issues} of machine learning and AI algorithms can also pose challenges in real-time gesture recognition applications.

\section{OmniRace Drone Control Scenario}

\subsection{System Overview} 
\begin{figure}[t]
\centerline{\includegraphics[width=0.475\textwidth]{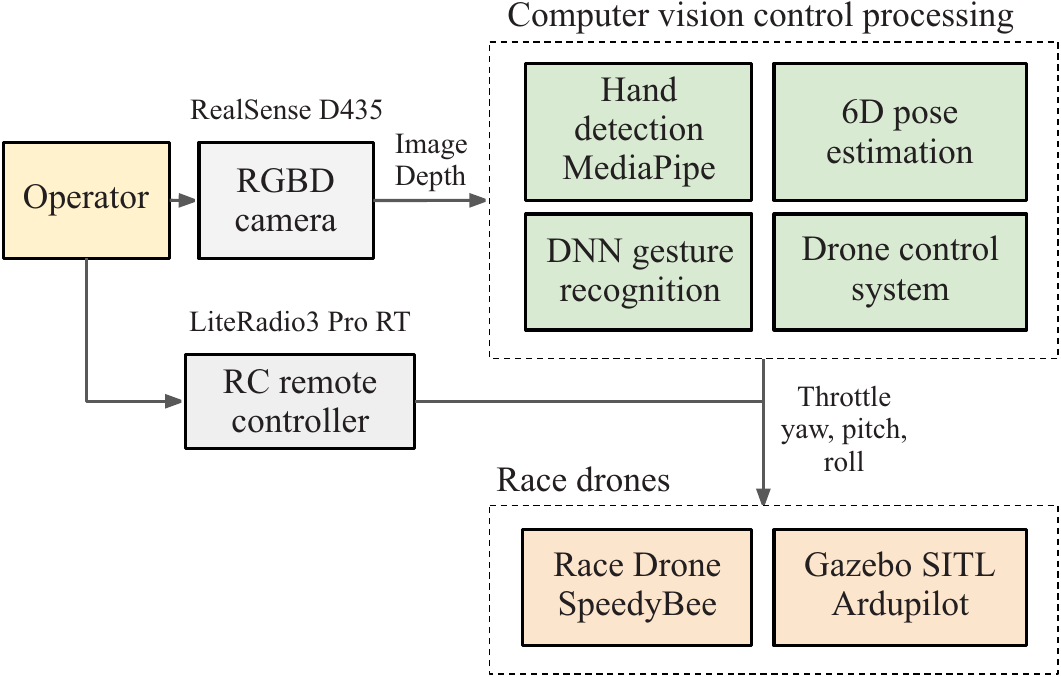}}
\caption{Algorithm pipeline for 6-DoF gesture-based drone race control.}
\label{fig:pipeline}
\end{figure}

The OmniRace system includes computer vision, with a hand recognition module utilizing the MediaPipe library, a pre-trained deep neural network (DNN) model for gesture recognition, and an algorithm to estimate the six degrees of freedom (6-DoF) of the operator's hand position and orientation. The information derived from the hand is then converted into a control signal for a drone within a software-in-the-loop (SITL) environment, such as a Gazebo simulation, or on an actual racing drone using a SpeedyBee F405 flight controller with ArduPilot autopilot firmware. The SITL gazebo plugin allows for fully simulated drone control, using the MAVROS interface, with the same behavior as a real drone. Input video frames for CV processing are received from a RealSense D435 camera at a resolution of 848 x 480 pixels and a frame rate of 30 fps, with RGB and depth frames aligned. A drone remote controller, LiteRadio3 Pro, is used for manual control and has been compared with the proposed gesture control method. All modules are shown in Fig. \ref{fig:pipeline}.

\subsection{DNN-based Gesture Recognition}

\begin{figure}[t]
\centerline{\includegraphics[width=0.36\textwidth]{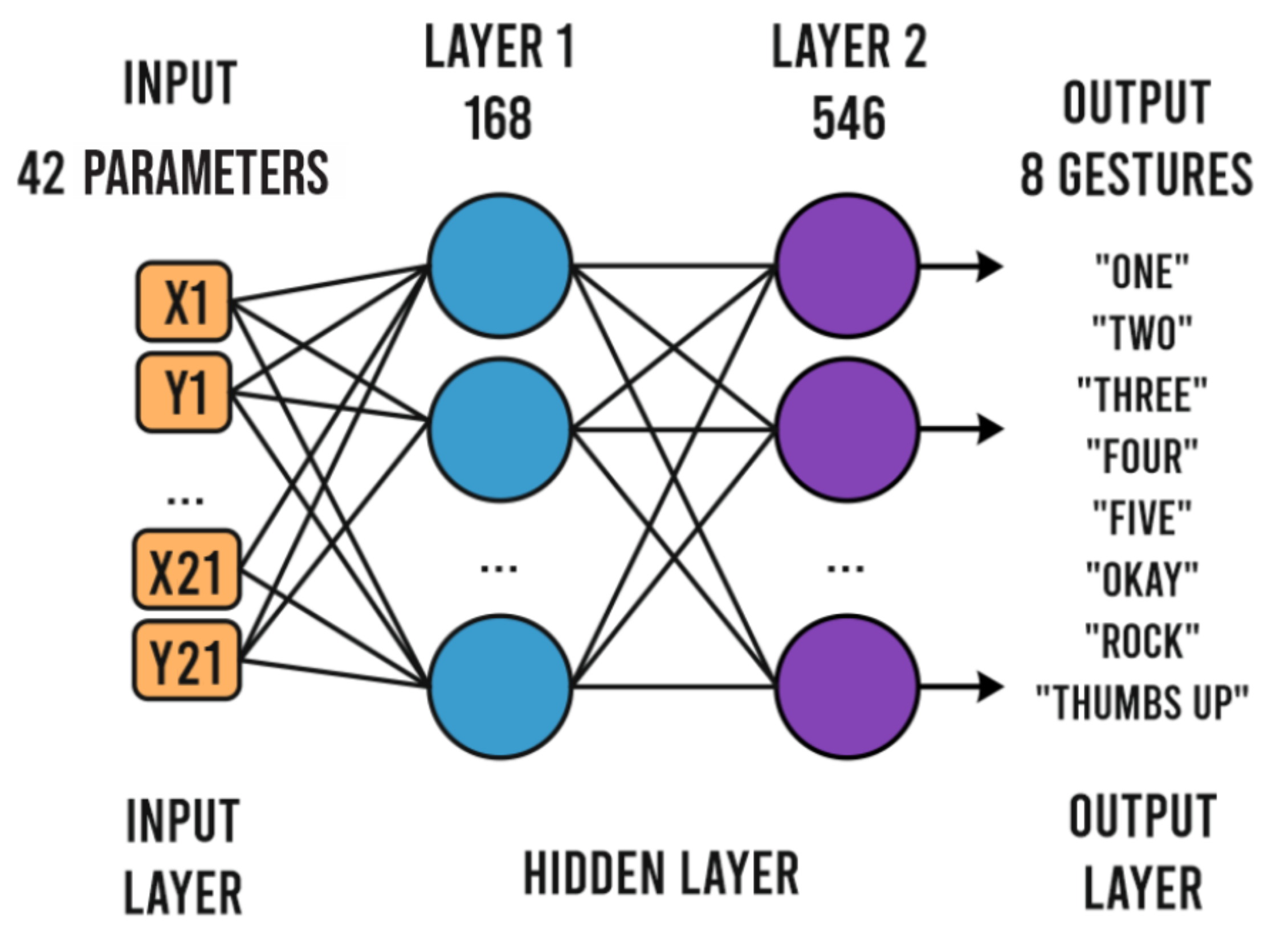}}
\caption{Structure of the DNN training model: 42 input parameters (x, y coordinates of one-hand landmarks), a hidden layer with 168 and 546 parameters, and an output layer with 8 parameters (representing gestures).}
\label{fig:dnn}
\end{figure}

\begin{figure}[t]
\centerline{\includegraphics[width=0.45\textwidth]{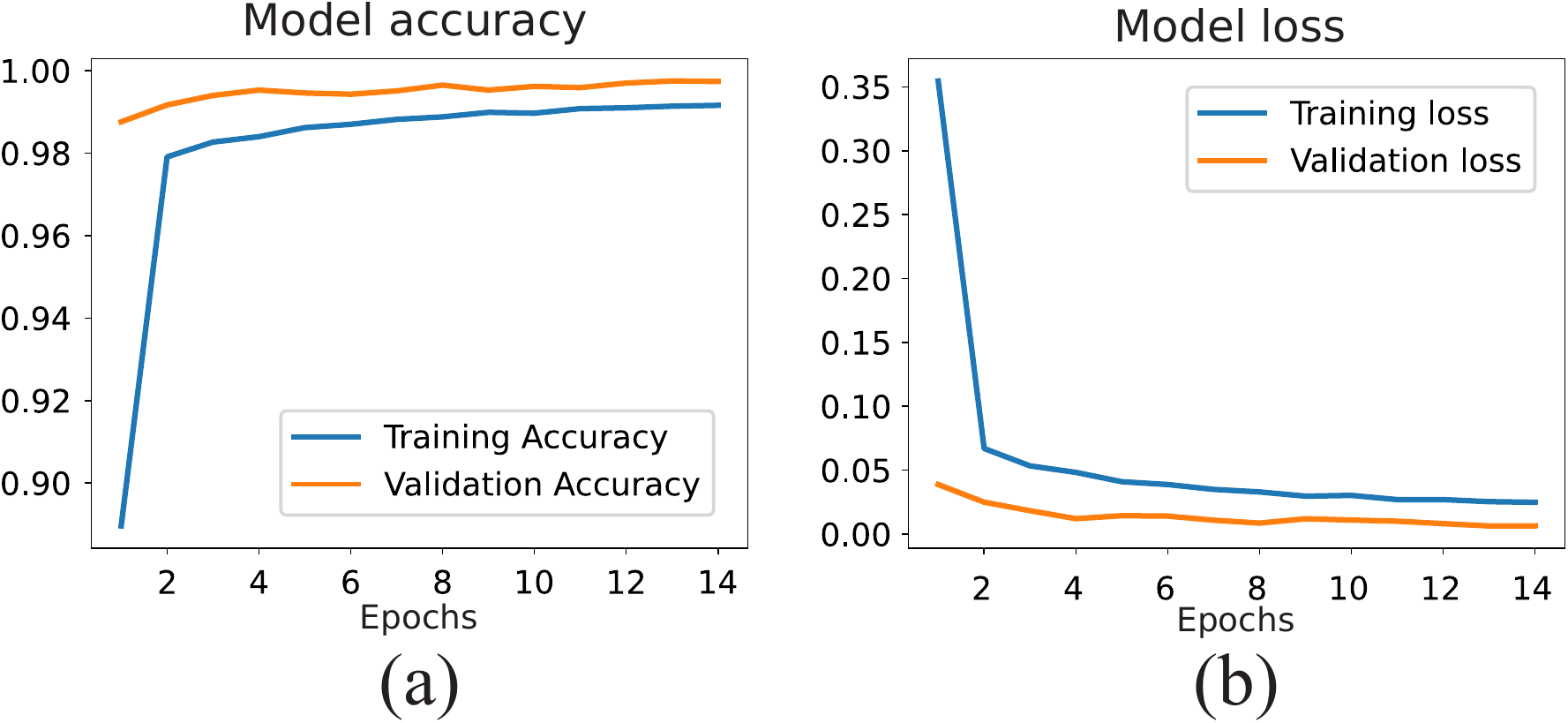}}
\caption{Graphs depicting accuracy (a) and loss (b) functions across multiple epochs. The yellow line represents validation data, while the blue line represents training data during the network learning process.}
\label{fig:learn_data}
\end{figure}

The MediaPipe Hand Landmarker detection \cite{mediapipe} was utilized as an advanced solution for tracking the movements of hands and fingers with high precision. The way it accomplishes this is by deriving 21 specific landmarks (X, Y) from each image frame of a hand. The gesture dataset was collected from five individuals and consists of 8,000 image arrays, each containing the coordinates of the 21 landmarks. This collected dataset additionally includes different hand positions for eight unique gestures. For each gesture, there are 1,000 unique point arrays that represent the specific positions of the hand, allowing for a highly detailed and precise representation of each gesture. OmniRace utilizes these eight gestures for high-level drone control (see Fig. \ref{fig:gesture_commands}), including: “one”, “two”, “three”, “four”, “five”, “okay”, “rock”, and “thumbs up”. The train-test split is 6400 to 1600, representing 80\% to 20\% data distribution. The DNN model, depicted in Fig. \ref{fig:dnn}, consists of three layers: input, hidden, and output. The input layer receives data features and the output layer delivers predictions of one of the eight gestures after processing through two hidden layers. The accuracy of gesture recognition on the test dataset is remarkably high, standing at 99.75\%. Epochs of learning are presented in Fig. \ref{fig:learn_data}. 

\subsection{6-DoF Hand Pose Estimation and Drone Control}

\begin{figure}[t]
\centerline{\includegraphics[width=0.35\textwidth]{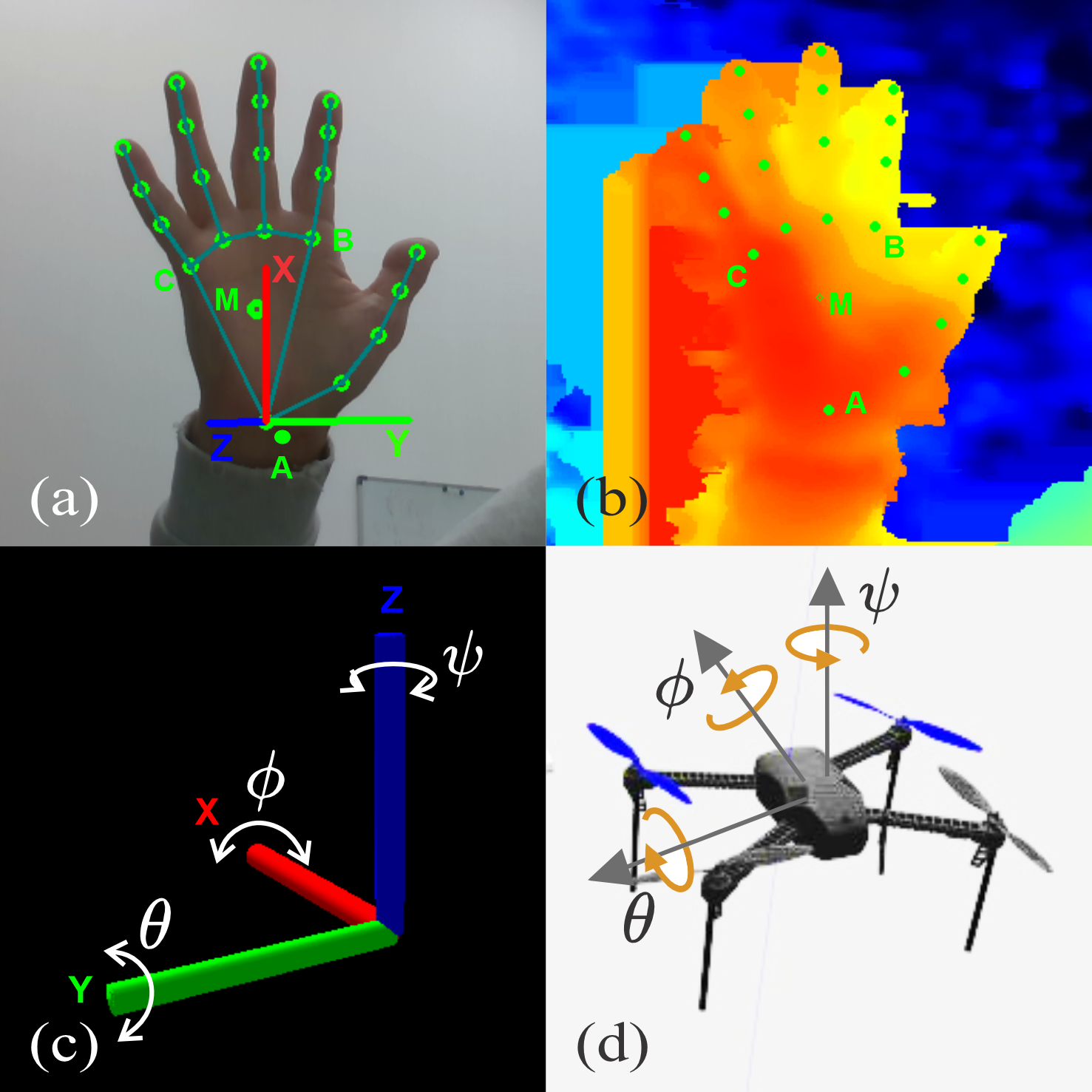}}
\caption{(a) Detection of 21 landmarks (x, y coordinates) of right-hand joint points. (b) Depth image of the hand with identified landmarks. (c) 3D hand pose and orientations calculated using ABCM points from color and depth images. (d) Transformed angles for drone orientation control.}
\label{fig:paln_orient}
\end{figure}

The process of determining the position and orientation of the hand involves calculating the relative translation of a hand in 3D space. The colored images from Fig. ~\ref{fig:paln_orient}(a) are used to obtain 2D positions of the 21 landmarks, while the depth frames are utilized to gather data for the 3D positioning of each hand point, as shown in Fig. ~\ref{fig:paln_orient}(b). We calculated the hand orientations in 3D by applying the method outlined in \cite{orient}. As shown in Fig. ~\ref{fig:paln_orient}(a), the hand normal vector $\overrightarrow{Z}$ is calculated as a vector product

\begin{equation}
\overrightarrow{Z} = \overrightarrow{AC} \times \overrightarrow{AB}.
 \label{eq:hand_vector}
\end{equation}

The hand normal vector $\overrightarrow{X}$ is calculated as a connection vector from the hand point $A$ to mean position $M$ of the hand point $A, B$, and $C$. The equation is shown as:
\begin{equation}
M = \frac 1 3 (A + B + C); \qquad \overrightarrow{X} = \overrightarrow{AM} 
 \label{eq: mean_point}
\end{equation}
where the vector $\overrightarrow{X}$ is also normalized to a unit vector. The calculation of the hand normal vector $\overrightarrow{Y}$ is ultimately done as the cross product of the normalized vectors $\overrightarrow{X}$ and $\overrightarrow{Z}$. Following this, the orientation of the hand is represented using quaternions $(q_\omega, q_x, q_y, q_z)$, as illustrated below:

\begin{equation}
    q_\omega = \frac{\sqrt{\overrightarrow{X_x}+ \overrightarrow{Y_y} + \overrightarrow{Z_z}}} {2} ;
\label{eq: quaternions}
\end{equation}

\begin{equation}
    q_x = \frac {\overrightarrow{Z_y} + \overrightarrow{Y_z}} {4q_\omega} ;
    q_y = \frac {\overrightarrow{X_z} + \overrightarrow{Z_x}} {4q_\omega} ;
    q_z = \frac {\overrightarrow{Y_x} + \overrightarrow{X_y}} {4q_\omega} .
\label{eq: quaternions}
\end{equation}

Then the quaternions are converted to Euler angles and represent as roll, pitch, yaw:
\begin{equation}
    \begin{bmatrix}
      \phi   \\  
      \theta \\  
      \psi  
    \end{bmatrix}
    =
    \begin{bmatrix}
      atan2( 2(q_\omega q_x + q_y q_z), 1 - 2(q_x^2 + q_y^2) )  \\  
      -\pi/2 + 2 atan2 (\sqrt{1 + 2 (Q)}, \sqrt{1 - 2 (Q} ) \\
      atan2( 2(q_\omega q_x + q_x q_y), 1 - 2(q_y^2 + q_z^2) ) \\
    \end{bmatrix}
    \label{eq: to_euler}
\end{equation}
where 
\begin{equation}
Q = q_\omega q_y - q_x q_z.
\end{equation}

Orientation angles, as shown in Fig. ~\ref{fig:paln_orient}(c), namely yaw, pitch, and roll, are transferred from the hand to the drone control system depicted in Fig. ~\ref{fig:paln_orient}(d). This allows the operator to set the drone's orientation at a low control level. The throttle is controlled by the depth value of point $M$. 

\begin{figure}[t]
\centerline{\includegraphics[width=0.325\textwidth]{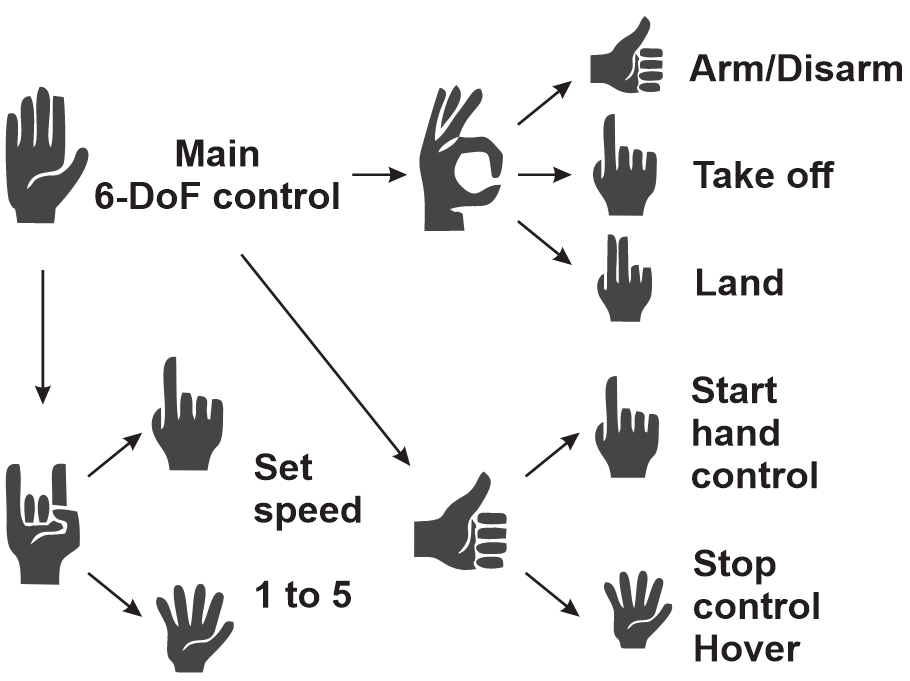}}
\caption{Combination of gesture commands used for high-level drone control.}
\label{fig:gesture_commands}
\end{figure}

The eight gestures recognized are utilized as high-level commands for the drone. Using a sequence of gestures allows the operator to send different commands. These commands include actions such as arming or disarming the drone, takeoff, landing, and setting the drone in flight mode or selecting a speed coefficient that is convenient for the user. The “five” is the main gesture to control a drone by orientation. When changing the gesture to select the speed or command, the drone switches to hover mode for safe flight. The set of gesture commands is shown in Fig. \ref{fig:gesture_commands}.

\section{Experimental Evaluation}

\subsection{Research Methodology}

In this study, we conducted a peer evaluation of the developed gesture interface by comparing it with a baseline remote controller. We invited 10 participants to engage in a drone race in the Gazebo simulation using two interfaces to control the drone flight. The recorded trajectories are shown in Fig. \ref{fig:traj_race}. Five participants did not have any prior experience in drone control and were categorized as the Beginner group, while the other five had some prior experience and were categorized as the Intermediate group. Participants started by controlling the drone initially using a remote controller, then by using the developed OmniRace gesture control interface. Short breaks were allowed between attempts and when switching control methods. Each participant was presented with a challenge to complete a race consisting of 10 gates, as shown in Fig. \ref{fig:teaser}(c). Before the race began, participants were given 30 minutes to familiarize themselves with both control methods. Subsequently, each participant performed three control flights. The racetrack was designed to test the ability to use all control inputs of a drone (yaw, pitch, roll, throttle), which are determined from gestures or remote control interfaces. Since not all the participants were professional drone race pilots, the drone in the simulation was controlled in position hold Ardupilot mode during the experiment. After the trials, each participant completed a three-part user survey, consisting of the unweighted NASA-TLX (scored out of 100 points on a 21-point scale) \cite{NASA}, SUS \cite{SUS}, and UEQ \cite{UEQ} questionnaires to assess the pragmatic and hedonic qualities of the gesture interface, as well as its overall performance.

\subsection{Experimental Results}
The results of the user study were processed both in one piece and separately for the Beginner and the Intermediate groups. Table \ref{tab:traj_results} presents the results of evaluation flight trajectories of all categories, and the results of the user questionnaire are represented in Table II.

\begin{figure}[t]
\centerline{\includegraphics[width=0.5\textwidth]{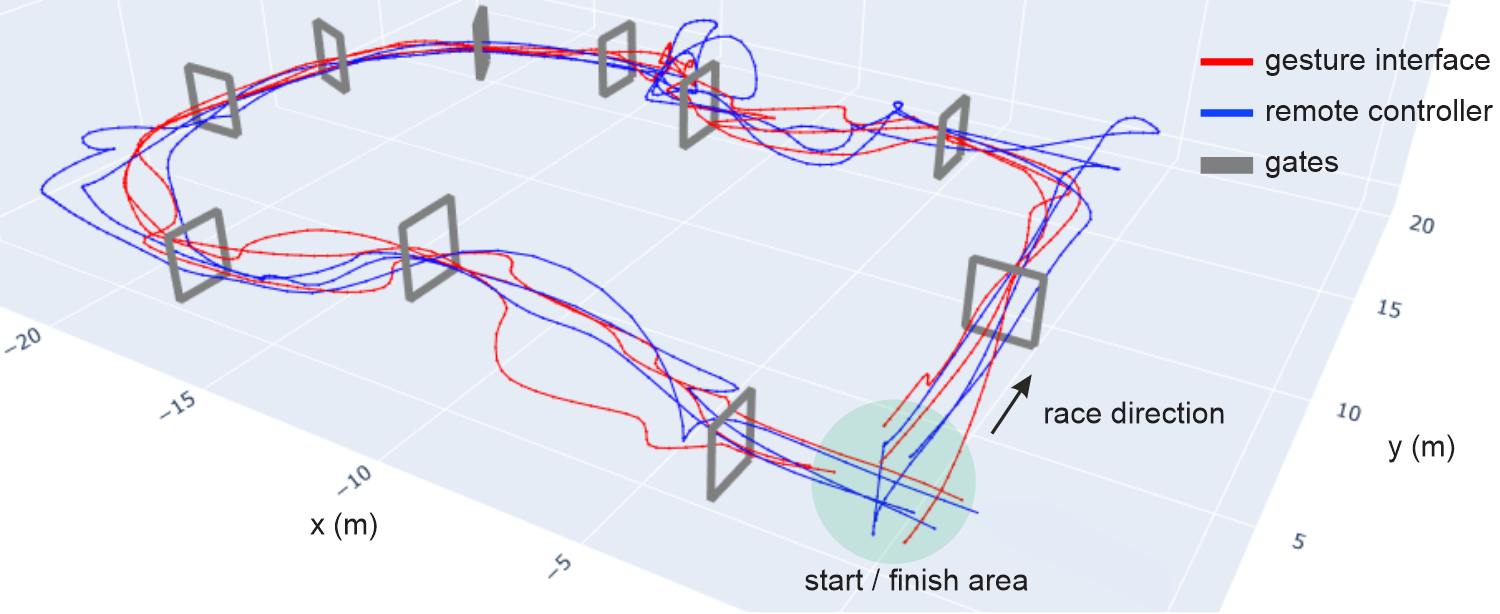}}
\caption{Set of recorded trajectories: gesture interface (red lines) versus remote controller (blue lines).}
\label{fig:traj_race}
\end{figure}

\begin{table}[b]
\caption{Experimental Results of Recorded Flight Trajectories by Gesture Interface and Remote Controller.} 
\label{tab:traj_results}
\centering
    \begin{tabular}{ |l|c|c|c| } 
        \hline
        \textbf{Participants} & \textbf{Metrics} & \begin{tabular}[c]{@{}c@{}} \textbf{Gesture}\\ \textbf{interface}\end{tabular} & \begin{tabular}[c]{@{}c@{}}\textbf{Remote}\\ \textbf{Controller}\end{tabular} \\
        \hline
        \multirow{3}{4em}{\centering Overall }
        & Time, s &  {{\textbf{103.25}}} & 129.18 \\ 
        & Length flight path, m & {{\textbf{83.70}}} & 102.90 \\ 
        & Average velocity, m/s & 0.92 & {{\textbf{1.13}}} \\ 
        \hline
        \multirow{3}{4em}{\centering Beginner level }
        & Time, s & {{\textbf{135.19}}} & 199.43 \\ 
        & Length flight path, m & {{\textbf{87.88}}} & 118.53 \\ 
        & Average velocity, m/s & {{\textbf{0.67}}} & 0.60 \\ 
        \hline
        \multirow{3}{4em}{\centering Intermediates level}
        & Time, s & 71.32 & {{\textbf{58.92}}} \\ 
        & Length flight path, m & {{\textbf{79.51}}} & 87.26 \\ 
        & Average velocity, m/s & 1.16 & {{\textbf{1.65}}} \\ 
        \hline
    \end{tabular}
\end{table}

The evaluation revealed that Beginners navigated the track on average one and a half times faster (135.19 s against 199.43 s) using the gesture control interface. In contrast, the Intermediates finished the race a bit faster by using the remote controller. Interestingly, the hand-controlled trajectories resulted in the shortest paths for all participants (mean 83.7 m) compared to those achieved using the remote control (mean 102.9 m). The average speed of the Beginners was the same across both control interfaces. The best time for gesture control was 54.8 s, with the shortest flight path of 71.34 m. The fastest time for remote control was 49.89 s, with the shortest flight path of 70.42 m. Also, the trajectories appeared smoother when controlled via gesture interface control than by a remote controller, as shown in Fig. \ref{fig:traj_race}.

\begin{table}[t]
    \caption{User Score of the Developed Interface with Unweighted NASA-TLX, UEQ, and SUS Scores, $p$ $\textless$ 0.05 Highlighted.}
    \label{tab:exp_results}
\centering
    \begin{tabular}{ |c|c|c|c| } 
        \hline
        \textbf{Score} & \begin{tabular}[c]{@{}c@{}}\textbf{Gesture}\\ \textbf{interface}\end{tabular} & \begin{tabular}[c]{@{}c@{}}\textbf{Remote}\\ \textbf{Controller}\end{tabular} & \begin{tabular}[c]{@{}c@{}}\textbf{Wilcoxon,} \\ \textbf{alpha of 0.05}\end{tabular} \\
        
        \hline
        \multicolumn{1}{|c}{NASA-TLX} & \multicolumn{3}{c|}{} \\
        \hline
        
        \multirow[t]{7}{*}{Mental Demand} & \textbf{31.5} & 56.5 &\textbf{V=7.5, p=0.04} \\
        Physical Demand & 29.5 & \textbf{19.0} & V=10.0, $p$=0.08 \\ 
        Temporal Demand & \textbf{32.0} & 49.0 & \textbf{V=6.0, $p$=0.04}\\ 
        Performance & \textbf{27.7} & 39.5 & V=14.5, $p$=0.62\\
        Effort & \textbf{32.0} & 48.0 &\textbf{V=1.5, $p$=0.01}\\
        Frustration & \textbf{16.5} & 43.5 &\textbf{V=2.5, $p$=0.02} \\
        Overall & 29.3 & 42.6 &\textbf{V=4.0, $p$=0.01} \\

        \hline
       
        \multicolumn{1}{|c}{UEQ}& \multicolumn{3}{c|}{} \\
        
        \hline
        
        \multirow[t]{7}{*} 
        {Attractiveness} & \textbf{1.57} & -0.35 & \textbf{V=0.0, $p$=0.002} \\
        Perspicuity & \textbf{1.55} & -0.55 & \textbf{V=0.0, $p$=0.007} \\ 
        Efficiency & -0.05 & \textbf{0.75} & V=11.0, $p$=0.17 \\ 
        Dependability & 0.65 & 0.20 & V=19.0, $p$=0.43 \\
        Stimulation & \textbf{1.40} & -0.13 & \textbf{V=0.0, $p$=0.012}\\
        Novelty & \textbf{1.73} & -1.80 & \textbf{V=1.0, $p$=0.003} \\
        Hedonic Quality & \textbf{1.56} & -0.96 & \textbf{V=1.0, $p$=0.003} \\
        Pragmatic Quality & \textbf{0.717} & 0.131 & V=12.0, $p$=0.13 \\ 
        
        \hline
        
        SUS  & 78.8 & 53.5 & N/A \\
        
        \hline
    \end{tabular}
    \label{tab:survey}
\end{table}

\begin{figure}[t]
\centerline{\includegraphics[width=0.45\textwidth]{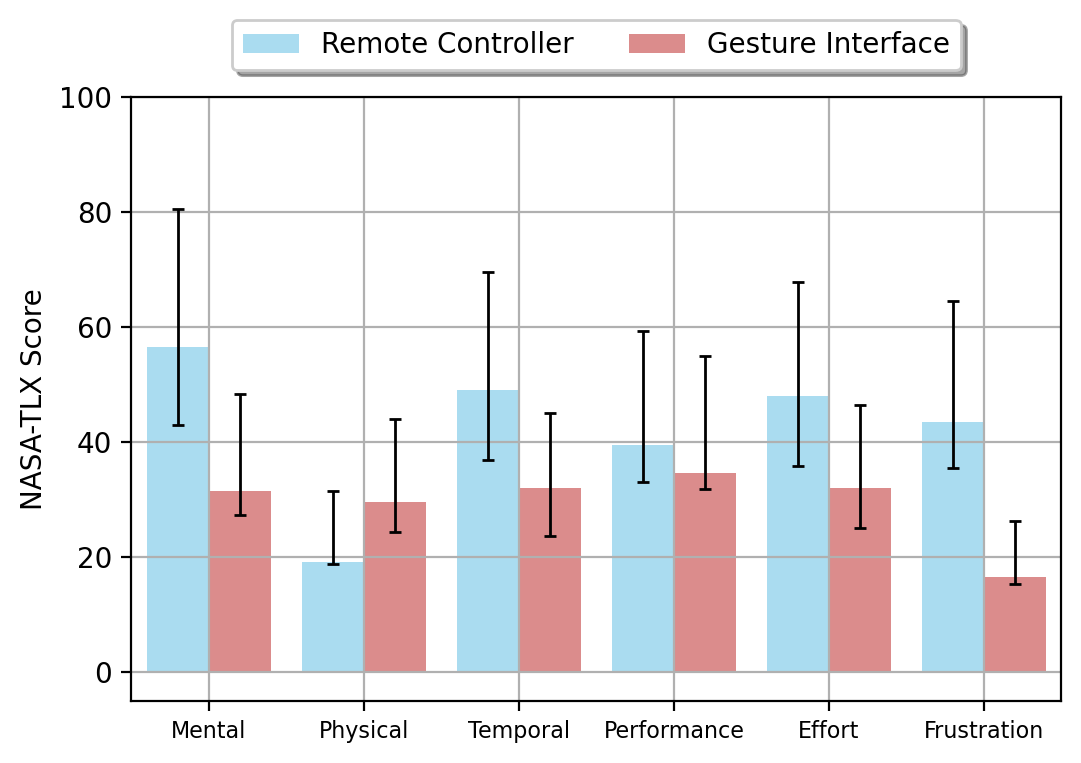}}
\caption{The visualization shows all dimensions of the NASA-TLX questionnaire. Black error bars denote 95\% confidence interval (CI).}
\label{fig:NASA}
\end{figure}

\begin{figure}[t]
\centerline{\includegraphics[width=0.45\textwidth]{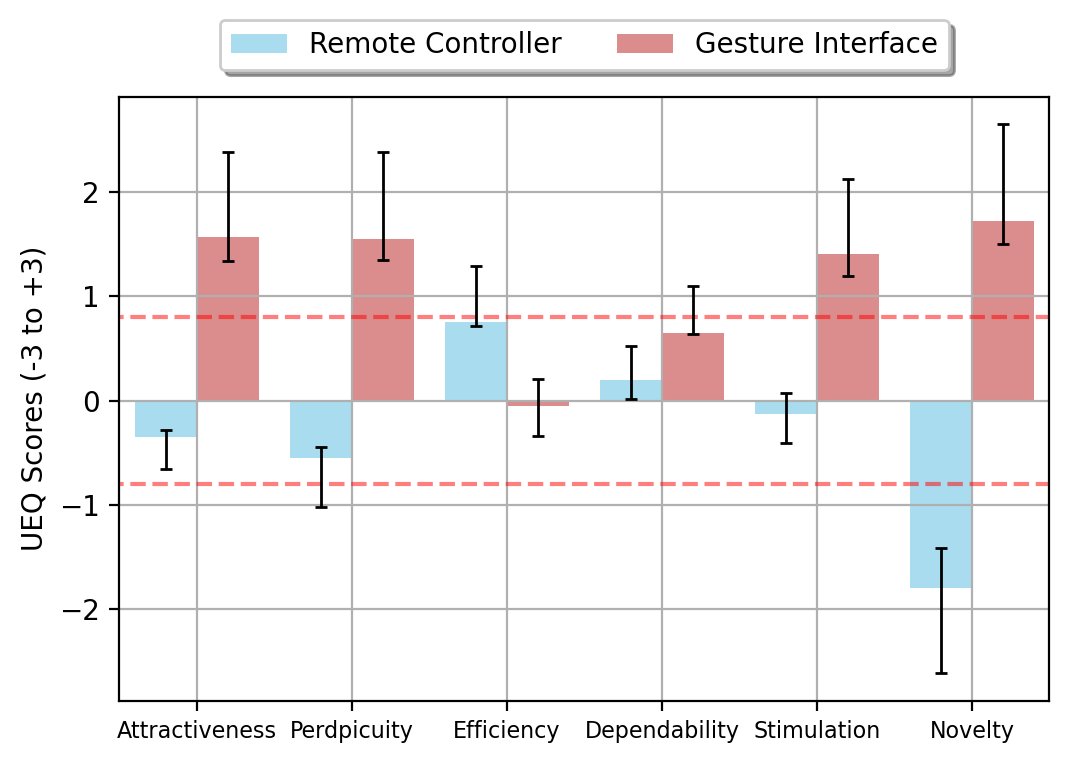}}
\caption{The visualization shows all dimensions of the UEQ questionnaire. Black error bars denote 95\% CI. Dotted red lines indicate UEQ's 0.8 threshold.}
\label{fig:UEQ}
\end{figure}

Given that the data were non-parametric, a Wilcoxon signed-rank test for paired samples was conducted. The NASA-TLX test, as represented in Fig. \ref{fig:NASA} revealed that participants‘ perceived workload differed significantly between the interfaces (V = 4.0, $p$ = $0.01 < 0.05$), resulting in a higher perceived workload in the case of the remote controller (M = 42.6, SD = 11.8) compared to the gesture interface (M = 29.3, SD = 9.8). Users expectantly evaluated the physical demand higher for the gesture interface (M = 29.5, SD = 19.2), however, statistical significance (V = 10.0, $p$ = 0.08) was not demonstrated for this metric. On the other hand, frustration (V = 2.5, $p$ = 0.02) was perceived by respondents lower for the gesture interface (M = 16.5, SD = 17.9) compared to the remote control (M = 43.5, SD = 26.8). We evaluated the SUS score as a standard metric for calculating the relative usability of the UAV control interfaces. Gesture Interface nearly reached an “excellent” SUS score of M = 78.8 (SD = 7.47, A- score), followed by Remote Controller which yielded an average SUS score of “OK” with M = 53.5 (SD = 17.00, D score). The UEQ dimensions for all drone control interfaces are visualized in Fig. \ref{fig:UEQ}.

In the UEQ evaluation, Gesture Interface received a positive user experience evaluation ($ > $ 0.8) for attractiveness, perspicuity, stimulation, and novelty. Most of the metrics for Remote Controller including attractiveness, efficiency, stimulation, and dependability, were scored as neutral (in the range of -0.8 and 0.8), while the highest score was received for efficiency. This result corresponds with the overall high performance of the experienced drone pilots with the controller, although the acquired data did not show a statistical difference in this metric. The results showed lower deviations in hedonic quality metrics, while pragmatic quality did not show a significant difference. One possible reason could be that all groups of users were satisfied more with their accomplishments when personally controlling the drone by hand rather than using the remote controller.

\section{Conclusions and Future Work}

In conclusion, this research introduces a novel approach to controlling racing drones using DNN-based 6-DoF pose estimation and hand gesture recognition. The experimental results demonstrate that OmniRace enabled users to navigate the UAV racing track significantly faster, with a 25.1\% reduction in completion time, and to shorten the length of the drone's path during testing by an average of 18\% compared to using the usual remote controller. Based on the evaluation results, we found that the gesture interface was perceived as less mentally demanding (NASA-TLX score of 31.5), significantly less frustrating (NASA-TLX score of 16.5), and superior in both hedonic and pragmatic qualities (1.56 and 0.717 UEQ scores) compared with the baseline remote controller. Additionally, experienced pilots were able to achieve a 9\% shorter trajectory when operating through OmniRace interface. 
This demonstrates the potential of the developed technology to transform drone racing and other applications where intuitive, real-time control of drones is required. 

For future research, we plan to conduct experiments with gesture-based control of real drones. We aim to address the limitations related to camera noise, frame rate, and image resolution to enhance control speed. Additionally, OmniRace could potentially be used to control drone swarms, a task not feasible with standard remote controllers.

\section*{Acknowledgements} 
Research reported in this publication was financially supported by the RSF grant No. 24-41-02039.

\bibliographystyle{IEEEtran}
\bibliography{reference} 
\end{document}